# Decision Making Using Probabilistic Inference Methods

**Ross D. Shachter**
Department of Engineering-Economic Systems
Stanford University
Stanford, CA 94305-4025, USA
(415) 723-4525, shachter@sumex-aim.stanford.edu

**Mark A. Peot**
Department of Engineering-Economic Systems
and
Rockwell International Science Center, Palo Alto Laboratory
444 High Street, Suite 400, Palo Alto, CA 94301, USA
(415) 325-7143, peot@rpal.rockwell.com

## Abstract

The analysis of decision making under uncertainty is closely related to the analysis of probabilistic inference. Indeed, much of the research into efficient methods for probabilistic inference in expert systems has been motivated by the fundamental normative arguments of decision theory. In this paper we show how the developments underlying those efficient methods can be applied immediately to decision problems. In addition to general approaches which need know nothing about the actual probabilistic inference method, we suggest some simple modifications to the clustering family of algorithms in order to efficiently incorporate decision making capabilities.

## 1 INTRODUCTION

Many important problems can be addressed using decision analysis, a probabilistic approach to making decisions under uncertainty. The long established method for evaluating these problems has been decision trees [Raiffa, 1968]. In recent years, the development of the influence diagram and reduction algorithms offered an efficient alternative for many classes of models [Howard and Matheson, 1984; Miller et al., 1976; Olmsted, 1983; Shachter, 1986]. Nonetheless, the basic problem is NP-hard, and both techniques become computationally infeasible for moderately sized problems. Thus, computational improvements to the solution of decision analysis problems can make a real difference in the successful application of decision analysis to real time systems.

The probabilistic inference problem is fundamentally linked to this decision problem. In fact, it is normative power of decision theory that has motivated many researchers to use a probabilistic approach to inference and learning. In recent years there have been significant advances in the development of algorithms for probabilistic inference in belief networks [Jensen et al., 1990a; Jensen et al., 1990b; Lauritzen and Spiegelhalter, 1988; Pearl, 1986]. In particular, the development of algorithms based on an undirected graphs has led to the current state of the art methods [Andersen et al., 1989]. In this paper, we show how this approach can be applied to the solution of decision problems.

The research in this area, although all closely related, can be divided into two main classes of algorithms. An asymmetric tree structure of decisions and observations can be used as a framework for solving a large number of closely related probabilistic inference problems [Cooper, 1988; Pearl, 1988]. Alternatively, a symmetric cross-sectional approach can be used to solve many of these problems simultaneously. That is the spirit behind the influence diagram algorithms based on node reductions [Ndilikilikesha, 1991; Olmsted, 1983; Shachter, 1986]. Recent work has shown, in a variant representation of the decision problem, the value of bringing the undirected graph to decision analysis [Shenoy, 1990; Shenoy, 1991a; Shenoy, 1991b]. In this paper, we show how these same concepts can be applied in an approach more consistent with current practice in probabilistic inference and decision analysis. The result is a simple method which allows the incorporation of decision making capabilities into all probabilistic inference systems, in a way which takes advantage of the some of the special structures in those systems.

Section 2 presents the influence diagram terms and notation. Section 3 shows how general belief network algorithms can be applied to decision making. Section 4 explores extensions and variations to the clustering family of algorithms in order to evaluate decisions, while Section 5 extends these results to dynamic programming problems. Finally, Section 6 presents conclusions.

## 2 MAKING DECISIONS

Influence diagrams are graphical representations for decision problems under uncertainty. In this section the components and notation of influence diagrams are introduced. The graphical structure of the influence diagram reveals conditional independence and the information needed to evaluate the decision problem. This is a cursory introduction and the reader is referred to the relevant literature for more information.

## 2.1 INFLUENCE DIAGRAMS

An **influence diagram** is a directed graph network representing a single decision maker's beliefs and preferences about a sequence of decisions to be made under uncertainty [Howard and Matheson, 1984]. The nodes in the influence diagram represent random variables (drawn as ovals), decisions (drawn as rectangles), and the criterion values for making decisions (drawn as a rounded rectangles). There are many names for influence diagrams containing only random variables, but we will simply call them **belief networks**. The parents of random variables and values are the conditioning variables for their distributions, while the parents of decisions represent those variables which will be observed before the decision must be made. Shaded random variable nodes, called **evidence** nodes, represent variables whose values have already been observed.

For example, consider the influence diagram shown in Figure 1a, which represents the decision whether to bring an umbrella to work. Our goal is to maximize our *Satisfaction* which depends on the *Weather* and on whether we *Bring Umbrella*. Our decision is *Bring Umbrella* and the key uncertainty is the *Weather*, which we won't observe until after we make our decision. We can learn about a weather *Forecast* before we make the decision; the *Forecast* depends on the *Weather*, or it wouldn't provide any useful information. A more complicated problem is represented by the influence diagram shown in Figure 1b. We have already observed the weather forecast in the *Newspaper*, which is dependent on the *Weather*. Now we get to choose which *TV Station* to watch for our weather *Forecast*, which now depends on both the *Weather* and the *TV Station*. We will know which *TV Station* we picked and its *Forecast* before we decide whether to *Bring Umbrella*.

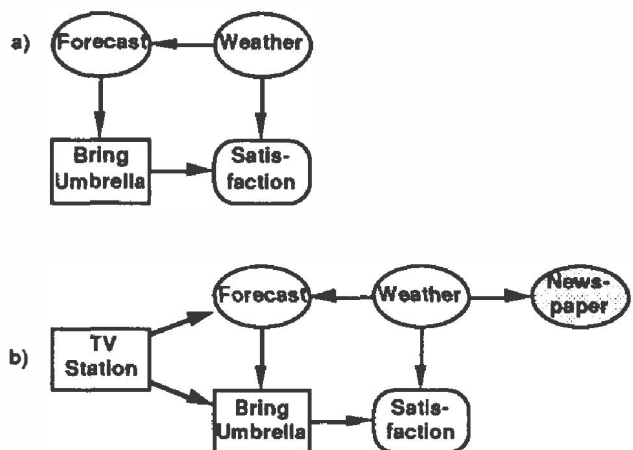

Figure 1. Examples of Influence Diagrams

The data in the influence diagram network is stored within the nodes. Each variable can take on some set of possible values: for random variables we will call these **outcomes**, for decisions we think of them as **alternatives**, and for the value these must be real numbers, so that we can make unambiguous choices. For each random variable, there is a conditional probability distribution giving the chances of different outcomes dependent on the outcomes of the variables parents. For value variables, there is a function, v( A ), giving the expected value as a function of its parents, called the **value attributes**, and denoted by the set A. Decision variables do not have a distribution, but once optimal choices can be determined the decision is replaced by a random variable, called the **optimal policy**, in which those choices can be indicated by a probability distribution or deterministic function. Finally, a full probability distribution is not needed for evidence nodes, since we already know which outcome has occurred. In these cases, a likelihood function is sufficient. Any child of an evidence node can reduce its distribution, since the evidence node cannot take on any other outcome [Shachter, 1989]. The influence diagram has been developed as a practical representation for a decision problem, and to that end that are several semantic restrictions, which are described in detail elsewhere [Howard, 1990; Howard and Matheson, 1984; Shachter, 1986].

There are several graphical/numerical operations called **reductions** which can be used to transform the influence diagram [Olmsted, 1983; Shachter, 1986; Shachter, 1988; Shachter, 1989]. These reductions are used to evaluate the influence diagram, determining the **optimal decision policies** and the **maximal expected value** or **MEV** of the decision problem. Within the diagram, it can be recognized that only a subset, $R^i \subseteq I^i$, of the non-evidence variables observed at the time of decision $D^i$ are needed to make the best choice. $R^i$ is the **relevant information** for $D^i$ and the optimal policy can be written as a function $d^{i*}$ from the possible values of $R^i$ to the alternative choices for $D^i$.

## 3 USING GENERAL PROBABILISTIC INFERENCE ALGORITHMS

In this section we are given a decision analysis problem represented by an influence diagram, and we wish to evaluate it using a general probabilistic inference system for belief networks. First we transform it into a belief network and then we show how to coordinate the analysis. The methods presented here are designed to be simple and straightforward for readability. There are some additional efficiencies to be gained by exploiting the methods described in Section 4.

### 3.1 INITIALIZATION

The influence diagram representation has a couple of elements not found in belief networks, namely decision and value nodes. These must be converted into probabilistic components before the decision problem can be solved using a probabilistic inference package. Before conversion, a simple linear-time algorithm can examine the influence diagram graph to determine which nodes can be omitted and, of those remaining, those for which probability distributions are not needed [Geiger et al., 1990; Shachter, 1988; Shachter, 1990].

Each decision node $D^i$ becomes a probabilistic node with parents $R^i$, the relevant information nodes for the decision. It has the decision alternatives as possible outcomes and will be given a probability distribution when the optimal policy has been determined. In the meantime, it should receive a uniform prior,

$$P\{ D^i = d_j \mid R^i \} = 1/N^i \text{ for } j = 1, \ldots, N^i,$$

where $N^i$ is the number of alternatives for decision $D^i$.

The value node is replaced by an "observed" probabilistic *Utility* node, U, with the value attributes, A, as parents and two outcomes, 0 and 1. The value function $v(a)$ should be rescaled into the interval [0, 1], defining a utility function, $u(a)$, the probability of outcome 1, to be

$$P\{ U = 1 \mid A = a \} = u(a) = \frac{v(a) - v_{min}}{v_{max} - v_{min}},$$

where $v_{min}$ and $v_{max}$ are the smallest and largest values of $v(a)$ [Raiffa, 1968; von Neumann and Morgenstern, 1947]. (Note: we can assume that $v_{max} > v_{min}$ without loss of generality, because if they were equal all alternatives would be equally attractive and there would be no real decision problem.)

### 3.2 QUERIES TO EVALUATE DECISIONS

Once a decision problem has been transformed into a belief network, a sequence of queries can be posed to a probabilistic inference algorithm so that it will compute the optimal policy for the decisions and the optimal value of objective function [Cooper, 1988; Pearl, 1988; Shachter, 1988].

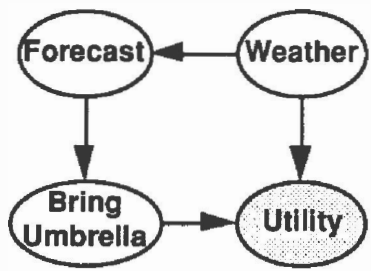

Figure 2. Belief Network Corresponding to Influence Diagram

First, consider a simple problem with only one decision and no evidence, as shown in Figure 1a. It will be transformed into the belief network shown in Figure 2. At this point, the observation for *Utility* has been entered, but the distribution of the decision policy, *Bring Umbrella*, has yet to be computed. To do so, we compute the joint distribution of the decision, $D^1$, and its relevant information, $R^1$, $P\{ D^1, R^1 \mid U=1 \}$. In this case, that would be $P\{$ *Bring Umbrella, Forecast* $\mid$ *Utility* =1 $\}$. Considering each possible case $r^1$ for $R^1$, define the function $d^{1*}(r^1)$ to be the maximizing alternative for $D^1$ as given by the following theorem. It may seem surprising that we select the most likely alternative given the best possible outcome, but that is equivalent to picking that alternative that maximizes the expected utility.

**Theorem 1.   Optimal Policies**
The optimal policy for decision $D^i$ is given by
$$d^{1*}(r^1) = \arg\max_{d^1} P\{ d^1, r^1 \mid U=1 \}.$$

Proof:
The joint distribution $P\{ U=1, d^1 \mid r^1 \}$ can be factored two ways, so
$$P\{ U=1, d^1 \mid r^1 \} = P\{ U=1 \mid d^1, r^1 \} P\{ d^1 \mid r^1 \}$$
$$= P\{ d^1 \mid U=1, r^1 \} P\{ U=1 \mid r^1 \}.$$
By construction, $P\{ d^1 \mid r^1 \}$ is the same for all possible values of $d^1$, thus
$$d^{1*}(r^1) = \arg\max_{d^1} E\{ u(A) \mid d^1, r^1 \}$$
$$= \arg\max_{d^1} P\{ U=1 \mid d^1, r^1 \}$$
$$= \arg\max_{d^1} \frac{P\{ d^1 \mid U=1, r^1 \} P\{ U=1 \mid r^1 \}}{P\{ d^1 \mid r^1 \}}$$
$$= \arg\max_{d^1} P\{ d^1 \mid U=1, r^1 \}$$
$$= \arg\max_{d^1} P\{ d^1 \mid U=1, r^1 \} P\{ r^1 \mid U=1 \}$$
$$= \arg\max_{d^1} P\{ d^1, r^1 \mid U=1 \}. \quad\#$$

If our only goal is to determine an optimal policy, then we are finished. Otherwise, we then enter a new distribution for $D^1$, with probability one for $P\{ D^1=d^1 \mid r^1 \}$ when $d^1 = d^{1*}(r^1)$ and zero otherwise. (In general, there might be more than one maximum corresponding to $r^1$; in that case, a "randomized strategy" could be entered provided that $\Sigma_{d^1} P\{ D^1=d^1 \mid r^1 \} = 1$.) Finally, we can now compute the maximal expected utility as
$$E\{ u(A) \mid d^{1*} \} = P\{ U=1 \mid d^{1*} \}.$$

In general, we might have evidence E=e, and a sequence of decisions, $D^1, \ldots, D^m$. In that case,
$$d^{m*}(r^m) = \arg\max_{d^m} P\{ d^m, r^m \mid U=1, E=e \},$$
and we can compute a new distribution for the optimal policy for $D^m$. Iterating backwards with $i = m-1, \ldots, 1$, we compute new distributions for $D^i$ based on
$$d^{i*}(r^i) = \arg\max_{d^i} P\{ d^i, r^i \mid U=1, E=e, d^{i+1*}, \ldots, d^{m*} \}.$$
Finally, we can compute the maximal expected utility as
$$E\{ u(A) \mid E=e, d^{i+1*}, \ldots, d^{m*} \}$$
$$= P\{ U=1 \mid E=e, d^{i+1*}, \ldots, d^{m*} \}.$$

## 4   CLUSTERING ALGORITHM MODIFICATIONS

In this section we adapt a clustering algorithm to evaluate decision problems under uncertainty. At first, for readability, this corresponds closely to the construction and queries applied to general probabilistic inference methods. The algorithm is extended by having it correspond more closely to the original influence diagram formulation, and refined through changes which increase its computational efficiency.

### 4.1   CLUSTER TREES

The clustering approach is particularly efficient for performing probabilistic inference on belief networks. Messages are passed between nodes in an undirected graph based on a belief network. By grouping variables together in nodes and allowing the same variable to appear in multiple nodes, a general, multiply-connected belief network can be represented by a tree. This tree organizes the factorization of the joint distribution for efficient computation [Jensen et al., 1990a; Jensen et al., 1990b; Lauritzen and Spiegelhalter, 1988; Shachter et al., 1991].

A set of variables is called a **cluster**. A tree of clusters is called a **cluster tree** (or join tree) for a belief network if each variable and its belief network parents appear together in at least one cluster, and if whenever a variable appears in two different clusters it appears in every cluster on the path between them.

Corresponding to each cluster is a **potential function**, which has the dimensions of the variables in the cluster, that is, there is a value of the potential function for each possible combination of values of the variables in the cluster. Messages are passed between neighboring clusters in a cluster tree; these messages have the dimensions of the variables in common between the two clusters. The conditional distribution for each variable is assigned to exactly one cluster, and by definition there will be at least one cluster with the proper dimensions. For each observation of evidence, there is a likelihood function which must also be assigned to exactly one cluster. The potential function for a cluster is initialized to the product of the conditional distributions and likelihood functions assigned to the cluster (or 1 if there is nothing assigned to the cluster).

Probabilistic inference is performed on a cluster tree by passing messages between clusters. The simplest way to organize this is through the **collect** operation, in which messages are sent from each terminal cluster node in the tree toward a particular cluster [Jensen et al., 1990b]. In this way, each cluster node and each arc are visited exactly once. Whenever a node is visited in this process it multiplies the messages it has received into its potential and then sends a message computed by summing out of its potential all of the variables which do not appear in its target neighbor. At the end, the potential at the collection cluster will be equal to the posterior joint distribution over that cluster's variables Z and the evidence E=e that was observed, P{ Z, E=e }. When additional collect operations are performed, care must be taken to ensure that each incoming message is multiplied into a cluster's potential function only once. For details on how this is managed see [Jensen et al., 1990a; Shachter et al., 1991].

### 4.2   UTILITY CLUSTERS

A cluster tree for a decision problem can be constructed from a belief network or directly from an influence diagram. Consider the influence diagram shown in Figure 1a, which is transformed into the belief network shown in Figure 2. It could be represented as any of the cluster trees shown in Figures 3a, 3b, and 3c. The cluster tree property requires that each variable appear in at least one cluster with all of its parents. In this case, *Weather* and *Bring Umbrella* must appear with *Utility*; *Forecast* must appear with *Bring Umbrella*; *Weather* must appear with *Forecast*; and *Weather* has to appear in some cluster. These conditions are satisfied by all three cluster trees.

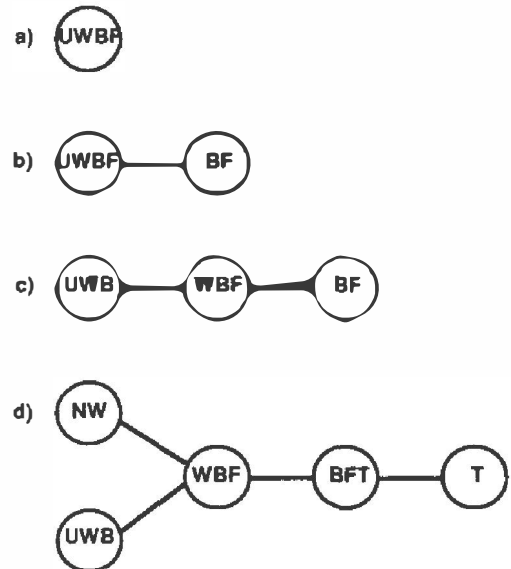

Figure 3. Cluster Trees Corresponding to Influence Diagrams

Consider instead the influence diagram shown in Figure 1b. It could be represented as the cluster tree shown in Figure 3d. The additional requirements are that *TV Station* and *Forecast* appear with *Bring Umbrella*, that *TV Station* appear in some cluster, and that *Newspaper* appear with *Weather*.

The general clustering method, with multiple decisions and evidence E=e, makes decisions in their reverse time order just like the probabilistic inference algorithm described in Section 3. Because each decision $D^i$ must appear in at least one cluster with $R^i$, we can use the potential functions for these clusters to determine the optimal policy. We collect to each of these clusters in reverse order and, letting Z represent the other variables present in the cluster for decision $D^i$,

$$d^{i*}(r^i) = \arg\max_{d^i} \Sigma_z \Psi_{D^i R^i Z}(d^i, r^i, z).$$

A probability distribution can now be entered for $P\{D^i \mid R^i\}$, equal to one for the optimal alternative and zero otherwise. (In general, there can be multiple optimal alternatives, so the probability distribution could represent a "randomized strategy.")

We can illustrate these operations with the cluster tree shown in Figure 3c corresponding to the influence diagram shown in Figure 1a. The utility function for *Utility* must be assigned to cluster $UWB$ and the conditional probability distribution for *Forecast* must be assigned to cluster $WBF$. The distributions for *Weather* could be assigned to either cluster but we will assign it to $WBF$. The utility is "observed" to have the outcome 1, so the likelihood function $f(u \mid w, b) = \begin{cases} 1 & \text{if } u=1 \\ 0 & \text{if } u=0 \end{cases}$ is also assigned to cluster $UWB$. Therefore the potential function for cluster $UWB$ is given by

$$\Psi_{UWB}(u, w, b) = \begin{cases} u(w, b) & \text{if } u=1 \\ 0 & \text{if } u=0 \end{cases}.$$

We will collect to the cluster $BF$ in order to determine the optimal choice for *Bring Umbrella*. First cluster $UWB$ sends a message to cluster $WBF$, obtained by summing out *Utility*. At that point, the potential function for cluster $WBF$ is given by

$$\Psi_{WBF}(w, b, f)$$
$$= P\{W=w\} P\{F=f \mid W=w\} \Sigma_u \Psi_{UWB}(u, w, b)$$
$$= P\{W=w\} P\{F=f \mid W=w\} u(w, b).$$

*Weather* is summed out to obtain the message from cluster $WBF$ to cluster $BF$. The potential function for cluster $BF$ becomes

$$\Psi_{BF}(b, f) = \Sigma_w \Psi_{WBF}(w, b, f).$$

This is precisely the information we need to make an optimal choice for *Bring Umbrella*,

$$b^*(f) = \arg\max_b \Psi_{BF}(b, f).$$

If we wish to go further, we can enter the probability distribution for *Bring Umbrella*,

$$P\{B=b \mid F=f\} = \begin{cases} 1 & \text{if } b=b^*(f) \\ 0 & \text{otherwise} \end{cases},$$

which can be multiplied into the potential function to obtain

$$\Psi_{BF}(b, f) \leftarrow \begin{cases} \Psi_{BF}(b, f) & \text{if } b=b^*(f) \\ 0 & \text{otherwise} \end{cases}.$$

We can sum out this potential to find the optimal expected utility,

$$E\{u(B, W) \mid b^*\} = \Sigma_{b,f} \Psi_{BF}(b, f).$$

### 4.3 VALUE CLUSTERS

The *Utility* variable, a rescaled value function, was introduced in Section 3 in order to allow expected value in a probabilistic setting. Within the clustering method, no such rescaling is necessary [Shenoy, 1990; Shenoy, 1991a; Shenoy, 1991b], but it is still desirable to maintain a special *Value* variable. When the influence diagram is converted to a belief network, the value node should be replaced by a *Value* node with two "outcomes" 0 and 1. Only now, the "probability" of those outcomes will be defined on the entire real line and the two "probabilities" will no longer sum to one. To make this distinction, we will use $V\{\}$ instead of $P\{\}$ and refer to this measure as a **valuation**. Thus

$$V\{V=v \mid A=a\} = \begin{cases} v(a) & \text{if } v=1 \\ 1 & \text{if } v=0 \end{cases}.$$

If *Value* is "observed" to be one, then all of the operations are unchanged from Section 4.2, except that the values are no longer rescaled. So, for example, we still have that

$$d^{i*}(r^i) = \arg\max_{d^i} \Sigma_z \Psi_{D^i R^i Z}(d^i, r^i, z).$$

On the other hand, if *Value* is not observed, it can stay in clusters yielding the similar result,

$$d^{i*}(r^i) = \arg\max_{d^i} \Sigma_z \Psi_{VD^i R^i Z}(v=1, d^i, r^i, z).$$

If *Value* is "observed" to be zero, then the potentials collected are now probabilities. The benefit of maintaining both cases comes when all of the other variables are summed from a cluster,

$$\Sigma_z \Psi_{VZ}(v, z) = \begin{cases} P\{E=e\} \text{ MEV} & \text{if } v=1 \\ P\{E=e\} & \text{if } v=0 \end{cases},$$

so that the maximal expected value, based on the decisions already made, is simply

$$\Sigma_z \Psi_{VZ}(v=1, z) / \Sigma_z \Psi_{VZ}(v=0, z).$$

Because *Value* is not a valid probabilistic variable, we should only send a message from a cluster with *Value* to one without it when we have "observed" *Value*, so that the message can be interpreted as either a valuation or a probability. It would be a mistake to sum over *Value*. As a result, most models should either have *Value* in many clusters, or just in one where it is "observed."

Another approach is that *Value* never needs to appear as an explicit variable. Instead of giving *Value* a valuation and observing it to take the value one, we can treat $v(a)$ like a likelihood function and assign it to a cluster containing the value attributes $A$. To compute the probabilities corresponding to $v=0$, we could either "forget" the likelihood function, or precompute $P\{E=e\}$. The tradeoff is between carrying *Value* around in clusters, thus doubling the size of the tables, or having to perform the computation twice.

From here on, we will assume that *Value* is used instead of *Utility*, since *Value* retains the units from the original influence diagram formulation.

### 4.4 ONE-DIRECTIONAL MESSAGE PASSING

The next development is to organize the cluster tree so that the message passing is as efficient as possible. Because multiple collect operations are performed and messages are passed throughout the network for each decision, the best case would be if every node and arc is visited only once in the course of the evaluation of all of the decisions. This is accomplished by establishing a direction for each arc, and passing a message exactly once in that one direction.

A cluster tree will be said to be **rooted** if there is a direction for each arc, every node has at most one child, and there is a unique childless node [Shenoy, 1990]. A rooted cluster tree will be called **one-directional** if it satisfies the following conditions:
1) the childless node has cluster $V$ or $\emptyset$ (the empty set); and
2) there is a directed path containing clusters for each decision $D^i$ such that:

   there is a cluster consisting of $D^i$, $R^i$, and optionally $V$, whose child cluster does not contain $D^i$; and
   these clusters appear in reverse decision order.

These conditions are quite restrictive, but it is easy to show that for every influence diagram there exists some one-directional rooted cluster trees. The reduction algorithm presented in Section 2.2 defines such trees, based on the sequence of clusters formed from the value node and its parents at every step.

**Theorem 2. Single Pass Evaluation**
When a rooted cluster tree is one-directional the entire decision problem can be evaluated in a single pass through the cluster tree.

Proof:
The full collect operation is being performed for the childless node in the tree, so all we need to show is that the decision policies based on partial collect operations will be correct. Consider the decision $D^i$ and the cluster $D^i R^i V$. Since the child cluster does not contain $D^i$, the missing message which would have come from that child could only have dimensions $R^i$ and $V$. Let such a message be $f(r^i, v)$. We know $f()$ is nonnegative, since negative factors only could come from the value function upstream of this cluster. (Otherwise, $D^i$ would be irrelevant to the value.) If we did a full collect operation to the cluster, we would obtain the posterior joint valuation, equal to $f$ times the potential (based on the distributions assigned to this cluster and the message incoming from its parent clusters):

$$V\{D^i, R^i, V, E=e\} = f(R^i, V) \, \Psi_{D^i R^i V}(D^i, R^i, V).$$

The policy $d^{i*}(r^i)$ conditioned on $R^i=r^i$ which is optimal with respect to $\Psi_{D^i R^i V}$,

$$d^{i*}(r^i) = \arg\max_{d^i} \Sigma_z \, \Psi_{D^i R^i Z}(d^i, r^i, v=1),$$

must therefore also be optimal with respect to $V\{D^i, R^i, V=1, E=e\}$.   #

Operations on a one-directional rooted cluster tree can be simplified even more. It is no longer necessary to enter a new distribution for $D^i$ after we determine the optimal policy and then sum it out; we can just maximize out the decision variable $D^i$ [Shenoy, 1990; Shenoy, 1991a; Shenoy, 1991b]. (This requires a one-directional tree, because the reverse operation is not well-defined.) When a variable is present in a sending cluster but not in the receiving cluster, it must be either maximized (for decisions variables) or summed (for other variables) out of the potential. If both types are present, then the decisions should be maximized before the others are summed.

Consider the cluster trees shown in Figure 4 corresponding to the influence diagrams shown in Figure 1. Each cluster tree is rooted and one-directional and the arcs have been marked with the operations to be performed on each cluster's potential to compute the message it sends to its child. The tree shown in Figure 4a has the *Value* variable present in every cluster, while the tree shown in Figure 4b has it present in just one cluster because it will be "observed" at $v=1$. The former tree corresponds to a sequence of reduction operations. In the latter tree, the message passed from cluster $BFT$ to cluster $T$ is computed by maximizing and then summing, $\Sigma_f \max_b \Psi_{BFT}(b, f, t)$.

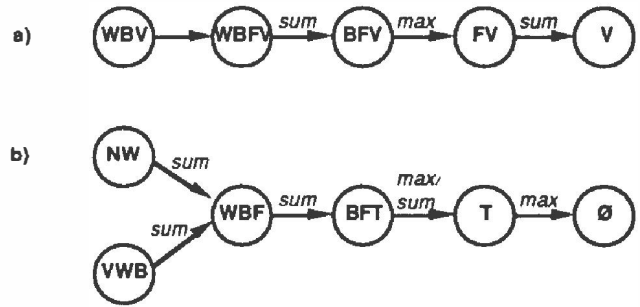

Figure 4. One-directional Rooted Cluster Trees

## 5. DYNAMIC PROGRAMMING

The original influence diagram representation has been extended to recognize the separable value function which allows for dynamic programming [Tatman, 1985; Tatman and Shachter, 1990]. The value function can be decomposed into a tree of sums and products, and this structure can be exploited by local computations. In this section we present some efficient analogs for these local computations in the modified cluster algorithm.

### 5.1 DYNAMIC INFLUENCE DIAGRAMS

Dynamic influence diagrams represent the separable structure of the value function as a tree of value nodes [Tatman, 1985; Tatman and Shachter, 1990]. In this paper we will consider only two structures, either a simple sum or a simple product. The value attributes, $A$, do not all condition the same value node directly, but instead the value is decomposed into multiple factors or terms which each have smaller value attribute sets. Consider the influence diagram shown in Figure 5. *Value* is either a sum or a product of *Value 1* and *Value 2*. This model, called a Markov decision process [Howard, 1960], is only drawn with two time periods, but it is clear how it could be drawn with any number n. We assume that *Value* is either the sum or the product of *Value 1*, ... , *Value n*.

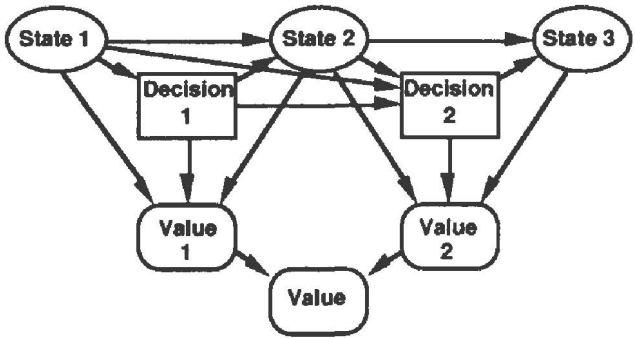

Figure 5. Markov Decision Process Dynamic Influence Diagram

These multiple value nodes allow the reduction operations to be applied with respect to a "local" value node rather than the "global" value node. For the dynamic influence diagram we must add one new reduction operation, the merger of local value nodes, in order to apply the reduction algorithm to the dynamic influence diagram [Tatman, 1985; Tatman and Shachter, 1990].

For example, starting with the diagram shown in Figure 5, *State 3* can be reduced into *Value 2*. We can now find an optimal policy for *Decision 2* with respect to *Value 2*, since all of the parents of *Value 2* are observed at the time of *Decision 2*. The policy node *Decision 2* can then be reduced into *Value 2*, but now we are stuck. At this point, *State 1* has a decision child, *Decision 1*'s value child depends on *State 2*, which is not observed at the time of *Decision 1*, and *State 2* has two value children. If we could merge *State 2*'s value children, then it could be reduced. Therefore, we merge *Value 1* and *Value 2* into *Value 1+2*, and reduce *State 2*, *Decision 1*, and *State 1* in that order.

When a policy is determined for *Decision 2*, *Value 2* only depends on *Decision 2* and *State 2*. Therefore, the optimal policy only depends on *State 2* even though *State 1* and *Decision 1* would be observed at the time of the decision. *State 2* is said to be a **Markov state** since it captures all of the information from the past necessary to make optimal decisions in the future. We can capture this relationship at the level of independence and relevance structures. First, let $W^i$ be the set of local value nodes which could be dependent on $D^i$ given the information available at the time of $D^i$, $I^i$. Now let the relevant information for $D^i$, $R^i$, be those variables in $I^i$ whose outcomes are needed to determine the optimal policy, $R^i = I^i \cap N_\Omega( W^i | D^i, I^i)$. There are linear-time algorithms to compute these sets [Geiger et al., 1990; Shachter, 1988; Shachter, 1990] We could also compute these sets by using the reduction algorithm graphically. If there were n time periods, then $W^i$ = { *Value i*, ... , *Value n* } and $R^i$ = { *State i* } for the Markov decision process.

The clustering algorithm is easily applied to dynamic influence diagrams in which the value structure is a simple product of nonnegative local values [Shenoy, 1990]. We can think of the value function as being decomposed into k factors, each with its own subset of the value attributes A.

The clustering algorithm is more complex to apply to dynamic influence diagrams in which the value structure is a simple sum, because the value summing operation does not correspond to the other operations on the cluster tree. We can think of the value function as decomposed into k terms, each of which must be maintained as a separate variable. Furthermore, it will be required that value variables are included in the decision clusters, and that the tree be one-directional.

The cluster tree for dynamic programming sums is constructed is two steps. First, we build a one-directional rooted cluster tree with the requirement that the cluster corresponding to decision $D^i$ now must consist of $D^i$, $R^i$, and the value variables $W^i$ that are the relevant for that decision. (Although these conditions are restrictive, they will be satisfied by any tree corresponding to reduction operations on the dynamic influence diagram.) Second, in any clusters with multiple value variables, replace those variables with a new variable representing their sum. A new operation must be defined to combine messages from clusters with different value variables.

## 6 CONCLUSIONS

We can get the full benefit of undirected graph probabilistic inference without having to abandon the influence diagram representation that decision analysts have found so useful for problem structuring and communication [Oliver and Smith, 1990]. The influence diagram is a natural representation for capturing the semantics of decision making with a minimum of clutter and confusion for the nonquantitative decision maker.

At the same time we get the performance dividends from undirected graph processing. If there is little inference in the problem, then the method presented here is essentially equivalent to influence diagram reductions. When there is complex evidence, it is not only more efficient, but also facilitates the recognition of opportunities for parallel processing. It also provides an opportunity to exploit many of the engineering advances incorporated into the best probabilistic inference algorithms [Andersen et al., 1989; Jensen et al., 1990a; Jensen et al., 1990b]. The results in this paper allow the systems developed for efficient probabilistic inference to incorporate efficient decision making as well. We believe that the absence of decision making in most probabilistic inference systems is most unfortunate and now we have shown how to correct that deficiency.

An alternative approach is based on valuation-based systems [Shenoy, 1990; Shenoy, 1991a; Shenoy, 1991b]. The particular power of this representation is its applicability to many different uncertainty calculi, but among those calculi, decision making is only well-defined for the Bayesian decision analysis paradigm, anyway. There can be a fixed computational advantage to this approach for some problems, but in essence the valuation and clustering approaches to decision making are both really incremental changes to the reduction algorithm, and all of these computational advantages can be obtained directly through a modification to the influence diagram [Ndilikilikesha, 1991].

There are many opportunities to extend and refine this research. In particular, there should be a simpler way to bring dynamic programming sums into the clustering algorithm. Also, these clustering methods can be easily applied to abduction problems: looking for the mostly likely outcome for a subset of variables given some evidence. Those problems would use the same operations presented here but with none of the order restrictions that complicate the evaluation of decisions. Finally, as always, we can benefit from better insight into the structuring of the cluster tree, since it can have such a significant impact on the algorithm's efficiency.

## 7 ACKNOWLEDGEMENTS

We benefitted greatly from the comments and suggestions of Stig Andersen, David Heckerman, Prakash Shenoy, the two anonymous referees, and a number of students in the EES department. This research was partially supported by the National Science Foundation through a Graduate Fellowship.